\documentclass[lettersize,journal]{IEEEtran}
\usepackage{amsmath,amsfonts}
\usepackage{algorithmic}
\usepackage{algorithm}
\usepackage{array}
\usepackage{listings}
\usepackage[caption=false,font=normalsize,labelfont=sf,textfont=sf]{subfig}
\usepackage{textcomp}
\usepackage{stfloats}
\usepackage{url}
\usepackage{verbatim}
\usepackage{graphicx}
\usepackage{cite}
\usepackage{hyperref}
\usepackage{color}
\usepackage{balance}
\usepackage{threeparttable}
\usepackage{booktabs}
\usepackage{xcolor}
\begin{document}

\title{Compositional Oil Spill Detection Based on Object Detector and Adapted Segment Anything Model from SAR Images}

\author{Wenhui Wu, Man Sing Wong, Xinyu Yu, Guoqiang Shi, Coco Yin Tung Kwok, and Kang Zou
\thanks{Manuscript received 15 January 2024; revised 3 March 2024; accepted 14 March 2024. Date of publication XXX; date of current version XXX. This work is funded by Environment and Conservation Fund. Any opinions, findings, conclusions or recommendations expressed in this material/event do not necessarily reflect the views of the Government of the Hong Kong Special Administrative Region and the Environment and Conservation Fund and the ‘Woo Wheelock Green Fund’ (for jointly funded project). M.S. Wong also thanks the support from the General Research Fund (project ID: 15609421 and 15603920), the Research Grants Council of Hong Kong. (\textit{Corresponding author: Man Sing Wong})}
\thanks{M.S. Wong is with the Department of Land Surveying and Geo-Informatics, The Hong Kong Polytechnic University, Kowloon, Hong Kong, and also with the  Research Institute for Sustainable Urban Development, The Hong Kong Polytechnic University, Hung Hom, Kowloon, Hong Kong (e-mail: Ls.charles@polyu.edu.hk).}
\thanks{W. Wu, X. Yu, G. Shi, C.Y.T. Kwok, and K. Zou are with the Department of Land Surveying and Geo-Informatics, The Hong Kong Polytechnic University, Kowloon, Hong Kong (e-mail: wenhui.wu@polyu.edu.hk; 19109597r@connect.polyu.hk; guoqiang.shi@polyu.edu.hk; yt-coco.kwok@connect.polyu.hk; welly.zou@polyu.edu.hk).}}

\markboth{Accepted to IEEE GEOSCIENCE AND REMOTE SENSING LETTERS}%
{Shell \MakeLowercase{\textit{et al.}}: A Sample Article Using IEEEtran.cls for IEEE Journals}

\IEEEpubid{0000--0000/00\$00.00~\copyright~2024 IEEE}

\maketitle

\begin{abstract}
  Semantic segmentation-based methods have attracted extensive attention in oil spill detection from SAR images. However, the existing approaches require a large number of finely annotated segmentation samples in the training stage. To alleviate this issue, we propose a composite oil spill detection framework, SAM-OIL, comprising an object detector (e.g., YOLOv8), an Adapted Segment Anything Model (SAM), and an Ordered Mask Fusion (OMF) module. SAM-OIL is the first application of the powerful SAM in oil spill detection. Specifically, the SAM-OIL strategy uses YOLOv8 to obtain the categories and bounding boxes of oil spill-related objects, then inputs bounding boxes into the Adapted SAM to retrieve category-agnostic masks, and finally adopts the OMF module to fuse the masks and categories. The Adapted SAM, combining a frozen SAM with a learnable Adapter module, can enhance SAM's ability to segment ambiguous objects. The OMF module, a parameter-free method, can effectively resolve pixel category conflicts within SAM. Experimental results demonstrate that SAM-OIL surpasses existing semantic segmentation-based oil spill detection methods, achieving mIoU of 69.52\%. The results also indicated that both OMF and Adapter modules can effectively improve the accuracy in SAM-OIL.
\end{abstract}


\section{Introduction}

\IEEEPARstart{S}{ynthetic} Aperture Radar (SAR) is one of the most efficient remote sensing tools for detecting oil spills \cite{brekke2005oil, al2020sensors}. One challenge in oil spill detection is to distinguish between oil spills and look-alikes, as both appear as dark spots in SAR imagery\cite{brekke2005oil, al2020sensors, alpers2017oil}. Dong \textit{et al.} \cite{Chronic2022} analyzed 563,705 Sentinel-1 images and found that about 90\% of oil slicks were located within 160 kilometers of shorelines, 21 among them are high-density slick belts which show similar to shipping routes. Therefore, shorelines and ships serve as crucial contextual information for oil spill detection and can assist in determining the source of oil spills \cite{Chronic2022, alpers2017oil}. Krestenitis \textit{et al.} \cite{krestenitis2019oil} developed a multi-class semantic segmentation dataset, which includes oil spill, look-alike, land, ship, and sea surface. In this letter, we cite this dataset as the Multimodal Data Fusion and Analytics (M4D) dataset, following the terminology of the proposing team's institution.

The traditional approach to detect oil spills on SAR images consists of three steps: dark spot segmentation, feature extraction, and distinguishing oil spills from look-alikes \cite{brekke2005oil, al2020sensors}. Firstly, dark spots on the SAR image are extracted through traditional segmentation methods. Then, various features such as geometric, contextual, texture, and physical features are extracted from these dark areas. Finally, a traditional classifier such as Support Vector Machine or Random Forest is used to distinguish oil spills from look-alikes \cite{brekke2008classifiers}. However, these methods have certain limitations, such as issues of under-segmentation or over-segmentation during the segmentation process, limited expressive capability of manually designed features, and the relatively weak discriminative ability of traditional classifiers in identifying oil spills. Therefore, there is still room for improvement in the accuracy of oil spill detection.

\begin{figure}[t]
  \centering
  \includegraphics[scale=0.5]{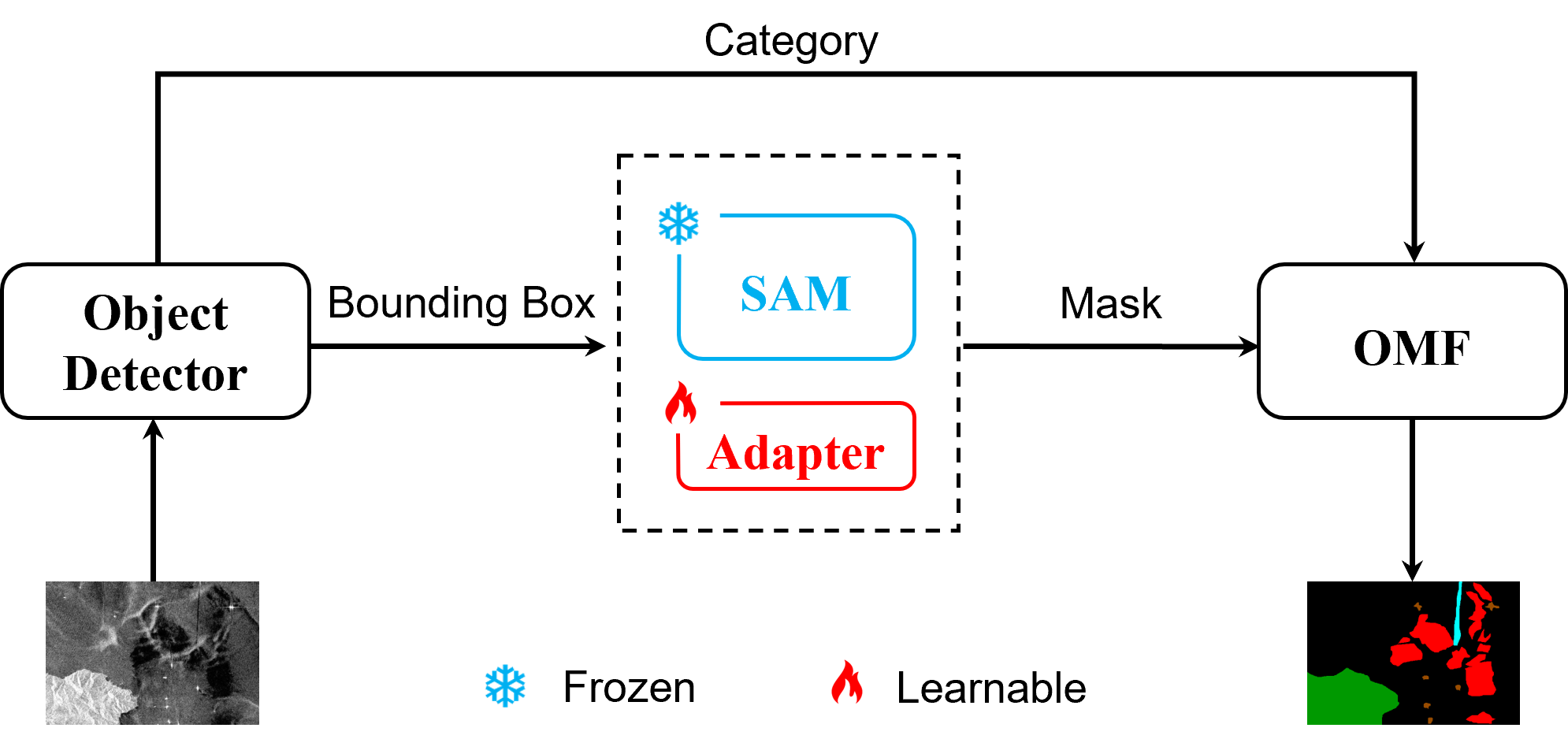}
  \caption{Architecture of the proposed method. The Adapted SAM is composed of a frozen SAM and a learnable Adapter module. SAM-OIL consists of an object detector, an Adapted SAM, and an OMF module.}
  \label{Fig1}
\end{figure}

\IEEEpubidadjcol

Deep learning approaches input with raw images and learn hierarchical feature representations from shallow visual features to deep semantic features from a large set of data. This process reduces the dependence on manually processing, demonstrating a strong capability to detect and capture complex patterns\cite{al2020sensors}. Therefore, deep learning-based methods have gained increasing attention in oil spill detection recently\cite{krestenitis2019oil, zhu2021oil, hasimoto2023ocean, ma2021oil}. Previous studies either directly used classical semantic segmentation models to detect oil spills, or developed tailor-made adaptive semantic segmentation models based on the characteristics of oil spills for the detection \cite{krestenitis2019oil, zhu2021oil}. In addition, the integration of polarimetric features in these models further enhances detection accuracy \cite{hasimoto2023ocean, ma2021oil}. Although these methods provided reasonably good detection performance, they required accurately annotated segmentation samples for training, which is time-consuming and resource-intensive for building the dataset.

To reduce the dependency on extensively annotated samples for semantic segmentation, we introduce SAM-OIL, a composite oil spill detection framework that includes an object detector, an Adapted Segment Anything Model (SAM), and an Ordered Mask Fusion (OMF) module. SAM represents the latest advancements in natural image segmentation, which offers impressive zero-shot segmentation performance by using input prompts such as points, bounding boxes, masks, and text \cite{Kirillov_2023_ICCV}. Among the types of prompts, those based on bounding boxes have led to the best segmentation results \cite{chen2023rsprompter}. Hence, we integrate an object detector (e.g., YOLOv8 \cite{mmyolo2022}) with SAM for oil spill detection. On the other hand, the outputs of SAM are category-agnostic binary masks, which may cause pixel category conflicts when masks are merged. To address this issue, we proposed the Ordered Mask Fusion (OMF) module, which merges masks according to a predefined category order. Furthermore, directly applying the SAM trained on natural images to remote sensing satellite imagery does not yield satisfactory results \cite{chen2023rsprompter}, especially for SAR images with blurred object boundaries. To bridge this gap, SAM needs to be adapted specifically for oil spill detection. HQ-SAM \cite{sam_hq} is the minimal adaptation of SAM, improving its segmentation precision by adding just one Adapter module, which accounts for no more than 0.5\% of SAM's parameters. Due to the simplicity and training efficiency of HQ-SAM, we integrate the Adapter module from HQ-SAM into SAM-OIL. The main contributions of this study are summarized as follows.
\begin{itemize}
  \item	We propose a compositional oil spill detection framework called SAM-OIL, which combines an advanced object detector (e.g., YOLOv8) with the SAM. To our best knowledge, this is the first time the SAM has been applied to oil spill detection. Furthermore, SAM-OIL framework can reduce the need for finely annotated samples.
  \item	Ordered Mask Fusion (OMF) module is proposed, which is a parameter-free method, to effectively resolve pixel category conflicts in SAM.
  \item	We introduce an Adapter module from HQ-SAM into the SAM-OIL, which utilizes masks from M4D dataset to train the Adapter, significantly enhancing SAM’s segmentation capabilities for objects with fuzzy boundaries.
  \item	Experimental results demonstrate that SAM-OIL achieves mIoU of 69.52\%, surpassing existing oil spill detection methods, and both OMF and Adapter can successfully improve the accuracy of SAM-OIL.
\end{itemize}

The rest of this letter is organized as follows. Section \ref{methodology} describes the methods proposed. Section \ref{experiments} performs a series of quantitative and qualitative analyses via experiments. Section \ref{conclusion} concludes this letter.

\section{Methodology} \label{methodology}
The architecture of the proposed method is given in Fig. \ref{Fig1}. SAM-OIL consists of an object detector (i.e., YOLOv8), an Adapted SAM, and an OMF module. We refer to the combination of YOLOv8 and frozen SAM as YOLOv8-SAM, which is considered as the baseline of our method.

\subsection{YOLOv8-SAM}
SAM consists of three components: an image encoder $\boldsymbol{\varPhi}_{enc-i}$, a prompt encoder $\boldsymbol{\varPhi}_{enc-p}$, and a lightweight mask decoder $\boldsymbol{\varPhi}_{dec-m}$. SAM takes an image $I$ and a set of prompts $P$ as inputs, which first uses $\boldsymbol{\varPhi}_{enc-i}$ to obtain image features $F_I$, and encodes the prompts $P$ into sparse prompt tokens $T_P$ through $\boldsymbol{\varPhi}_{enc-p}$. Then, $T_P$ and some learnable output tokens $T_O$ are concatenated, and finally input into $\boldsymbol{\varPhi}_{dec-m}$ along with $F_I$ for attention-based feature interaction, generating category-agnostic binary masks $M$. This process can be described as:
\begin{equation}	\label{eq1}
  \begin{cases}
    F_I &=\boldsymbol{\varPhi}_{enc-i}(I) \\
    T_P &=\boldsymbol{\varPhi}_{enc-p}(P)\\
    M &=\boldsymbol{\varPhi}_{dec-m}(F_I, Concat(T_O, T_P))
  \end{cases}
\end{equation}

YOLOv8-SAM is composed of an object detector and a frozen SAM. The procedure is as follows: 1) The object detector $\boldsymbol{\varPhi}_{det}$ obtains the object categories $C_{bbox}$, bounding boxes $P_{bbox}$, and classification score $S_{bbox}$. $P_{bbox}$ is filtered based on the defined threshold $S_{threshold}$ for $S_{bbox}$. 2) The image $I$ and a set of prompts $P_{bbox}$ are input into SAM, to obtain prompt tokens $T_{bbox}$ and masks $M_{bbox}$. 3) The $M_{bbox}$ and the corresponding $C_{bbox}$, are randomly merged to obtain the final masks $M_{semantic}$. Considering the effectiveness of the YOLO series models, we choose the latest YOLOv8 as the object detector. The complete process is described as follows:
\begin{equation}	\label{eq2}
  \begin{cases}
    C_{bbox}, P_{bbox} &=
    \begin{cases}
      \boldsymbol{\varPhi}_{det}(I)\\
      S_{bbox} > S_{threshold}
    \end{cases}\\
    F_I &=\boldsymbol{\varPhi}_{enc-i}(I) \\
    T_{bbox} &=\boldsymbol{\varPhi}_{enc-p}(P_{bbox})\\
    M_{bbox} &=\boldsymbol{\varPhi}_{dec-m}(F_I, Concat(T_O, T_{bbox}))\\
    M_{semantic} &=Random(C_{bbox}, M_{bbox})
  \end{cases}
\end{equation}

\begin{algorithm}[H]
	\caption{Pseudocode of OMF in a Numpy-like style.}
	\label{alg1}
	\definecolor{codeblue}{rgb}{0.25,0.5,0.5}
	\lstset{
		backgroundcolor=\color{white},
		basicstyle=\fontsize{7.2pt}{7.2pt}\ttfamily\selectfont,
		columns=fullflexible,
		breaklines=true,
		captionpos=b,
		commentstyle=\fontsize{7.2pt}{7.2pt}\color{codeblue},
		keywordstyle=\fontsize{7.2pt}{7.2pt},
	}
  \begin{lstlisting}[language=python]
    # Detector outputs categories, SAM outputs masks.
    classes, bboxes, scores = Detector(img)
    masks = SAM(img, bboxes)
    
    # Pre-defined category fusion order.
    order = ["ship", "land", "oil_spill", "look-alike"]
    
    # Sort masks according to the 'order'.
    zipped = zip(classes, masks)
    sorted_masks = sorted(zipped, order)
    
    # Fuse the masks.
    result = zeros((img_h, img_w, 1))
    for mask in sorted_masks:
        result[result==0] = mask[result==0]
  \end{lstlisting}
\end{algorithm}
 
\subsection{Ordered Mask Fusion module}
The Ordered Mask Fusion (OMF) module is proposed to address pixel category conflicts encountered in SAM's mask fusion by using a specific order for masks, as detailed in Algorithm \ref{alg1}. The order determines the categorization of ambiguous pixels. For instance, if the order is "ship, land, oil spill, look-alike", and a pixel is identified as both "ship" and "land", it will be classified as "ship".

\subsection{Adapted SAM}

We adopt HQ-SAM \cite{sam_hq} as the Adapted SAM because of its simplicity and training efficiency. The HQ-SAM is composed of a frozen SAM and a learnable Adapter module. During the training process, all the pre-trained SAM parameters are frozen, while only the Adapter module is updated. Adapter module increases $\leq$ 0.5\% of SAM's parameters, thereby ensuring the module's efficiency in training \cite{sam_hq}. The HQ-SAM was initially developed to enhance SAM's segmentation accuracy for intricate structures in natural images, while the HQ-SAM used in this study aims to improve SAM's ability to segment blurry boundaries in SAR imagery. The Adapter module is trained for specific tasks like oil spill detection. Thus, using SAM-OIL with this module may change or decrease accuracy in other tasks, but without affecting SAM's generalization. Additionally, the Adapter module can be retrained for different applications.

HQ-SAM was constructed by introducing the fusion of HQ-Output Token and deep-shallow feature into SAM. Firstly, a learnable HQ-Output Token $T_{HQ}$ is designed, which is input into SAM's mask decoder $\boldsymbol{\varPhi}_{dec-m}$ together with the original  prompts $T_P$ and output token $T_O$. Unlike the original output token $T_O$, the HQ-Output Token $T_{HQ}$ and its related Multi-layer Perceptron (MLP) layer are trained to predict high-quality segmentation masks. Secondly, by fusing the early feature $F_{early}$ and late feature $F_I$ of its image encoder, it uses both global semantic context and local fine-grained features. We refer to the combination of the HQ-Output Token, three-layer MLP, and a small feature fusion block as the Adapter module of HQ-SAM. The combination of the Adapter module and the original SAM decoder $\boldsymbol{\varPhi}_{dec-m}$ forms a new decoder $\boldsymbol{\varPhi}_{dec-m}'$. In summary, the SAM-OIL is composed of an object detector, HQ-SAM, and an OMF module, of which the whole process can be described as:
\begin{equation}	\label{eq3}
  \begin{cases}
    C_{bbox}, P_{bbox} &=
    \begin{cases}
      \boldsymbol{\varPhi}_{det}(I)\\
      S_{bbox} > S_{threshold}
    \end{cases}\\
    F_{early}, F_I &=\boldsymbol{\varPhi}_{enc-i}(I) \\
    T_{bbox} &=\boldsymbol{\varPhi}_{enc-p}(P_{bbox})\\
    T &= Concat(T_{HQ}, T_O, T_{bbox})\\
    M_{bbox} &=\boldsymbol{\varPhi}_{dec-m}'(F_{early}, F_I, T)\\
    M_{semantic} &=\boldsymbol{\varPhi}_{OMF}(C_{bbox}, M_{bbox})
  \end{cases}
\end{equation} 

\section{Experiments} \label{experiments}

\subsection{Experimental Setups}

\textit{1) Dataset:} The M4D dataset \cite{krestenitis2019oil} is used in this study, which is composed of five categories: oil spill, look-alike, land, ship, and sea surface. It includes a total of 1,112 SAR images, each with a dimension of 650$\times$1250 pixels. The images were divided into training and testing sets, with 1,002 images (90\%) for training and 110 images (10\%) for testing. 

\textit{2) Evaluation Metrics:} We use Intersection over Union (IoU), mean Intersection over Union (mIoU) and mF1 (mean F1-score) for evaluation, commonly applied in semantic segmentation. IoU measures the accuracy of individual classes, while mIoU and mF1 assess overall accuracy. 

\textit{3) Configuration Details:} All the experiments were conducted on two NVIDIA TITAN RTX GPUs (48G memory), with the operating system of Ubuntu 18.04.6 LTS. The object detector used was YOLOv8-X in MMYOLO \cite{mmyolo2022}, configured for 1,000 training epochs (note, early stopping strategy was employed) and batch size of 6, with all other parameters kept at MMYOLO's default settings. We also used a pre-trained model on the COCO dataset to initialize YOLOv8-X parameters \cite{mmyolo2022}. To maintain consistency with HQ-SAM's training scheme, the M4D dataset was refined by creating a binary mask for each object, without including any additional class information.  These binary masks were then used to train the HQ-SAM Adapter module. The training involved 120 epochs, using the original image size (650$\times$1250) and bounding box prompts. The ViT-H \cite{Kirillov_2023_ICCV} model was employed as the image encoder for the SAM. The remaining parameters followed the HQ-SAM's default configuration.

\textit{4) Comparative experiment setup details:}  Inspired by \cite{krestenitis2019oil}, we applied several classic semantic segmentation methods implemented in MMSegmentation \cite{mmseg2020} to the M4D dataset, including U-Net, PSPNet, UPerNet, DeepLabV3+, and OCRNet, with each model's backbone detailed in Table \ref{tableI}. All comparative methods were conducted on the same hardware configuration as SAM-OIL, with training epochs set at 300, batch size adjusted according to model size, and all other training hyperparameters following MMSegmentation's default settings.

\subsection{Comparative Experiments}
Table \ref{tableI} compares the experimental results of our methods with other methods. SAM-OIL achieved 69.52\% mIoU, higher than DeepLabV3+ (67.59\%), by 1.93\%. It also achieved the best results in the "Oil Spill" and "Ship" categories. Notably, in the "Ship" category, SAM-OIL reached IoU of 52.55\%, higher than DeepLabV3+ (36.99\%), by 15.56\%. This indicates SAM-OIL, incorporating an object detector, has significant advantages in detecting small objects. Moreover, SAM-OIL outperforms our baseline model, i.e.,YOLOv8-SAM, across all categories, demonstrating the effectiveness of the OMF and Adapter modules used (see the Ablation Study section for detailed analysis). However, in the "look-alike" category, SAM-OIL's IoU was 55.60\%, lagging behind the best model, OCRNet (63.53\%), by 7.93\%. It still requires improvement.

\begin{figure}[!h]
	\centering
	\includegraphics[scale=0.08]{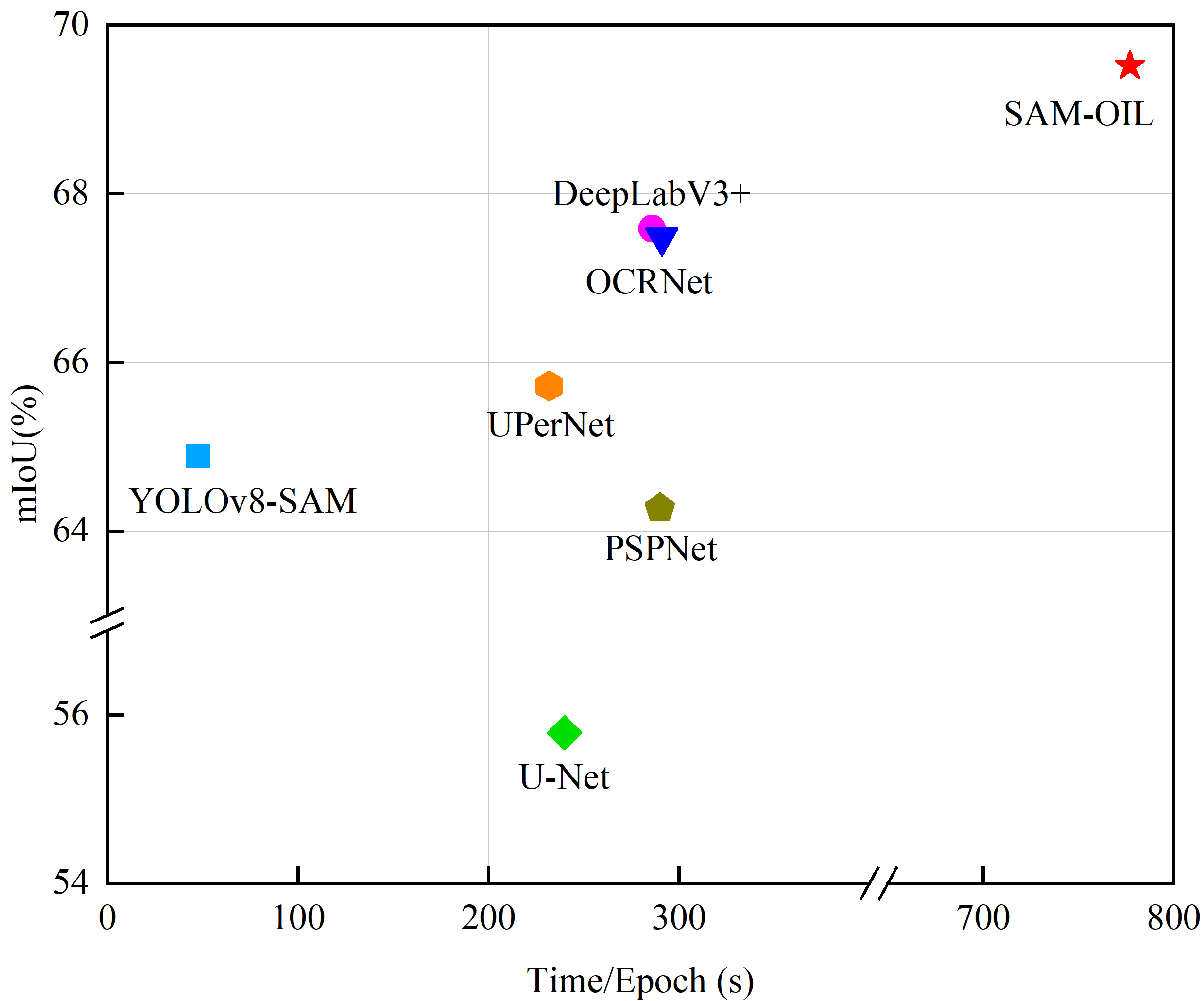}
	\caption{The epoch time for each model.}
	\label{Fig2}
\end{figure}

Table \ref{tableI} lists each model's parameters, showing that SAM-OIL has more than the others, which means it uses more computational resources and is slower in training and inference. Fig.\ref{Fig2} shows that YOLOv8-SAM achieves the fastest training speed per epoch since it only trains YOLOv8, while SAM-OIL trains both YOLOv8 and the Adapter, leading to longer training times. However, training YOLOv8 and the Adapter is more efficient than training the entire SAM. A lighter SAM version, like EfficientSAM \cite{xiong2023efficientsam}, could improve SAM-OIL’s training and inference speed.

\begin{table*}
	\caption{Quantitative Performance Comparison Between The Proposed Method and The Semantic Segmentation Methods on the M4D Dataset.}
	\centering
	\begin{tabular}{ccccccccccc}
		\toprule
		Methods& Backbone &Params.(M)& Sea Surface & Oil Spill & Look-alike & Ship & Land & mIoU &mF1\\
		\midrule
		U-Net&/&\textbf{28.99}&92.21&43.92&31.04&21.07&90.74&55.79&66.86\\
		PSPNet&ResNet50&46.60&96.80&44.37&59.77&27.61&92.78&64.27&74.84\\
    UPerNet&ResNet50&64.04&96.51&48.67&58.17&31.38&93.89&65.72&76.37\\
    DeepLabV3+&ResNet50&41.22&97.01&49.09&61.79&36.99&93.06&67.59&78.22\\
    OCRNet&ResNet50&70.35&\textbf{97.03}&50.07&\textbf{63.53}&32.65&\textbf{94.11}&67.48&77.82\\
    \midrule
    YOLOv8-SAM&YOLOv8-X+ViT-H&68.16+2446&94.34&41.84&48.15&52.48&87.65&64.89&76.67\\
    SAM-OIL&YOLOv8-X+ViT-H&68.16+2452.10&96.05&\textbf{51.60}&55.60&\textbf{52.55}&91.81&\textbf{69.52}&\textbf{80.43}\\
		\bottomrule
	\end{tabular}
    \begin{tablenotes}
      \centering
      \footnotesize
      \item * All results are described using percentages (\%). The accuracy of each category is evaluated using IoU. The best score is marked in bold.
    \end{tablenotes}
	\label{tableI}
\end{table*}

\subsection{Ablation Study}

\begin{table*}[ht]
  \begin{minipage}[t]{0.48\textwidth}
  \centering
  \caption{The influence of object detector on the accuracy.}
  \begin{tabular}{cc}
  \toprule
  Methods & mIoU \\
  \midrule
  YOLOv8-SAM & 64.89 \\
  YOLOv8-SAM (gt\_box) & 77.09 \\
  SAM-OIL & 69.52 \\
  SAM-OIL (gt\_box) & \textbf{80.35} \\
  \bottomrule
  \end{tabular}
  \label{tableII}
  \end{minipage}%
  \hfill
  \begin{minipage}[t]{0.52\textwidth}
  \centering
  \caption{The influence of OMF and Adapter on the accuracy.}
  \begin{tabular}{cccc}
  \toprule
  Methods & OMF & Adapter & mIoU \\
  \midrule
  YOLOv8-SAM & & & 64.89 \\
  YOLOv8-SAM + OMF & \checkmark & & 65.82 \\
  YOLOv8-SAM + Adapter & & \checkmark & 67.97 \\
  SAM-OIL (YOLOv8-SAM + OMF + Adapter) & \checkmark & \checkmark & \textbf{69.52} \\
  \bottomrule
  \end{tabular}
  \label{tableIII}
  \end{minipage}
  \end{table*}
  
\begin{table*}
	\caption{The influence of fusion orders of object masks in OMF on the accuracy.}
	\centering
	\begin{tabular}{ccccccc}
		\toprule
		Orders& Sea Surface & Oil Spill & Look-alike & Ship & Land & mIoU\\
		\midrule
    Random&96.05&48.34&54.87&50.10&90.47&67.97\\
    Look-alike, Oil Spill, Ship, Land&96.05&48.31&54.84&41.78&90.41&66.28\\
    Look-alike, Oil Spill, Land, Ship&96.05&48.31&54.84&41.78&90.41&66.28\\
    Look-alike, Land, Oil Spill, Ship&96.05&48.31&54.84&41.78&90.41&66.28\\
    Ship, Oil Spill, Land, Look-alike&96.05&51.60&55.60&52.55&91.81&69.52\\
    Ship, Land, Oil Spill, Look-alike&96.05&51.60&55.60&52.55&91.81&69.52\\
    Land, Ship, Oil Spill, Look-alike&96.05&51.60&55.60&52.55&91.81&69.52\\
	  \bottomrule
  \end{tabular}
	\label{tableIV}
\end{table*}

\textit{1) Detector accuracy:} To assess the impact of detector performance on the accuracy of SAM-OIL, ground truth bounding boxes (gt\_box) are input into SAM and Adapted SAM respectively. In Table \ref{tableII}, YOLOv8-SAM (gt\_box) surpasses YOLOv8-SAM by 12.2\%, and SAM-OIL (gt\_box) surpasses SAM-OIL by 10.83\%. It demonstrates that the detector significantly impacts the accuracy within the SAM-OIL framework.

\begin{figure}[!h]
	\centering
	\includegraphics[scale=0.15]{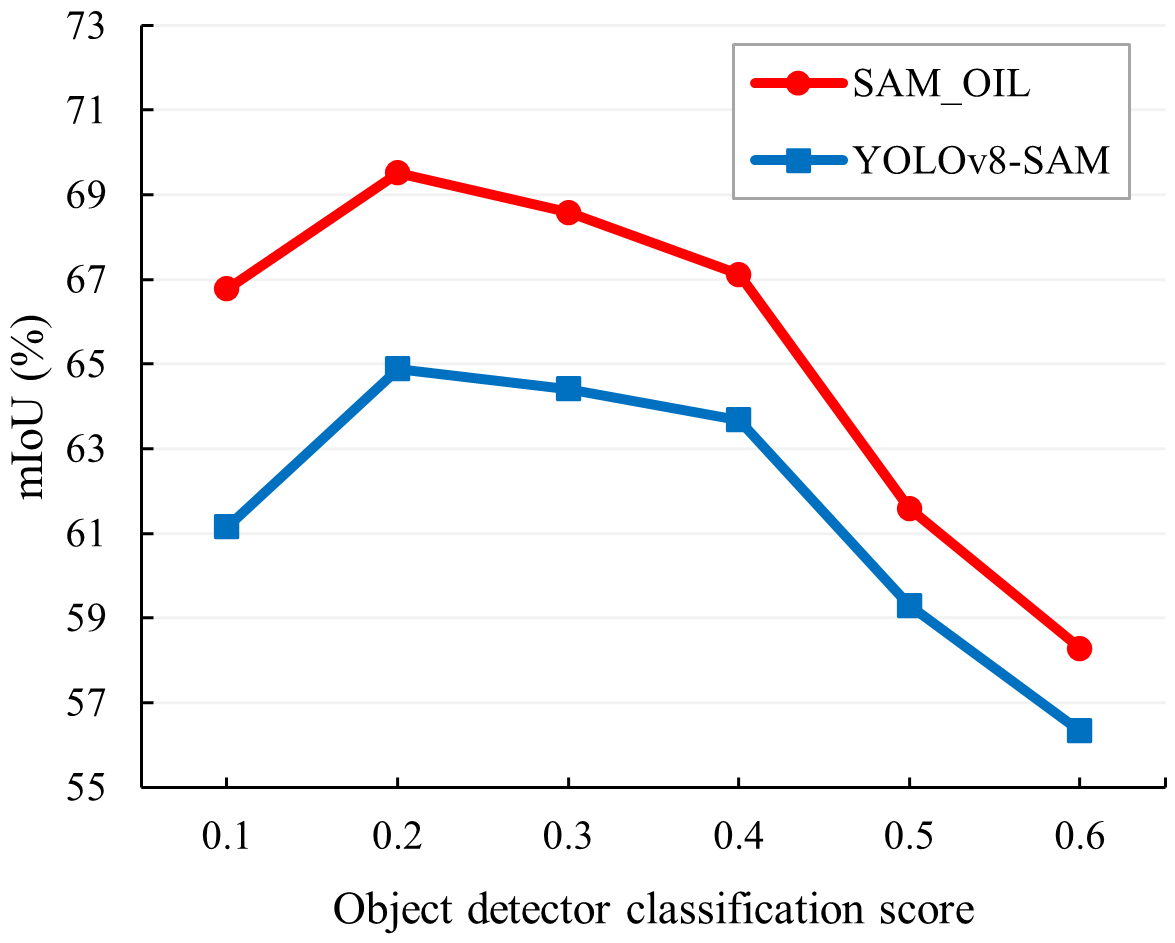}
	\caption{The influence of object detector classification scores on the accuracy.}
	\label{Fig3}
\end{figure}

\textit{2) Detector classification scores:} Besides bounding boxes and their corresponding categories, the detector also outputs classification scores, which can help determine if boxes should be ignored. Fig. \ref{Fig3} shows that different score thresholds impact accuracy; with a 0.2 score, YOLO-SAM and SAM-OIL reach peak accuracy. Increasing the score significantly lowers accuracy as higher scores may filter out correct detections.

\textit{3) OMF:} Table \ref{tableIII} illustrates that YOLOv8-SAM+OMF outperforms YOLOv8-SAM by 0.93\%, and SAM-OIL exceeds YOLOv8-SAM+Adapter by 1.55\%. Notably, OMF operates without training parameters. YOLOv8-SAM+OMF reaches 65.82\% mIoU and needs just box-level annotated samples for training, proving the potential of SAM-based composite methods in reducing the workload of sample annotation.

\textit{4) Adapter:} In Table \ref{tableIII}, YOLOv8-SAM+Adapter surpasses YOLOv8-SAM by 3.08\%, and SAM-OIL surpasses YOLOv8-SAM+OMF by 3.7\%. Furthermore, comparing YOLOv8-SAM in Table \ref{tableI} with Random in Table \ref{tableIV} shows that the Adapter module increases IoU by 6.5\% for "oil spill" and 6.72\% for "look-alike". This emphasizes the role of Adapter in enhancing SAM's ability to segment objects with ambiguous boundaries.

\textit{5) Order of OMF:} SAM outputs masks in four categories: ship, land, oil spill, and look-alike, with 24 possible permutations plus a random order, totaling 25. Table \ref{tableIV} shows accuracy for random, worst, and best orders, emphasizing a 3.24\% accuracy gap between the best and worst. This suggests order significantly impacts accuracy, advising higher priority for "ship" and "land" due to their effect on classification accuracy and size in images, and lower for "look-alike" due to its proximity to other categories and lower accuracy, to improve overall performance.

\subsection{Qualitative Analysis}

\begin{figure}[!h]
  \centering
  \includegraphics[scale=0.5]{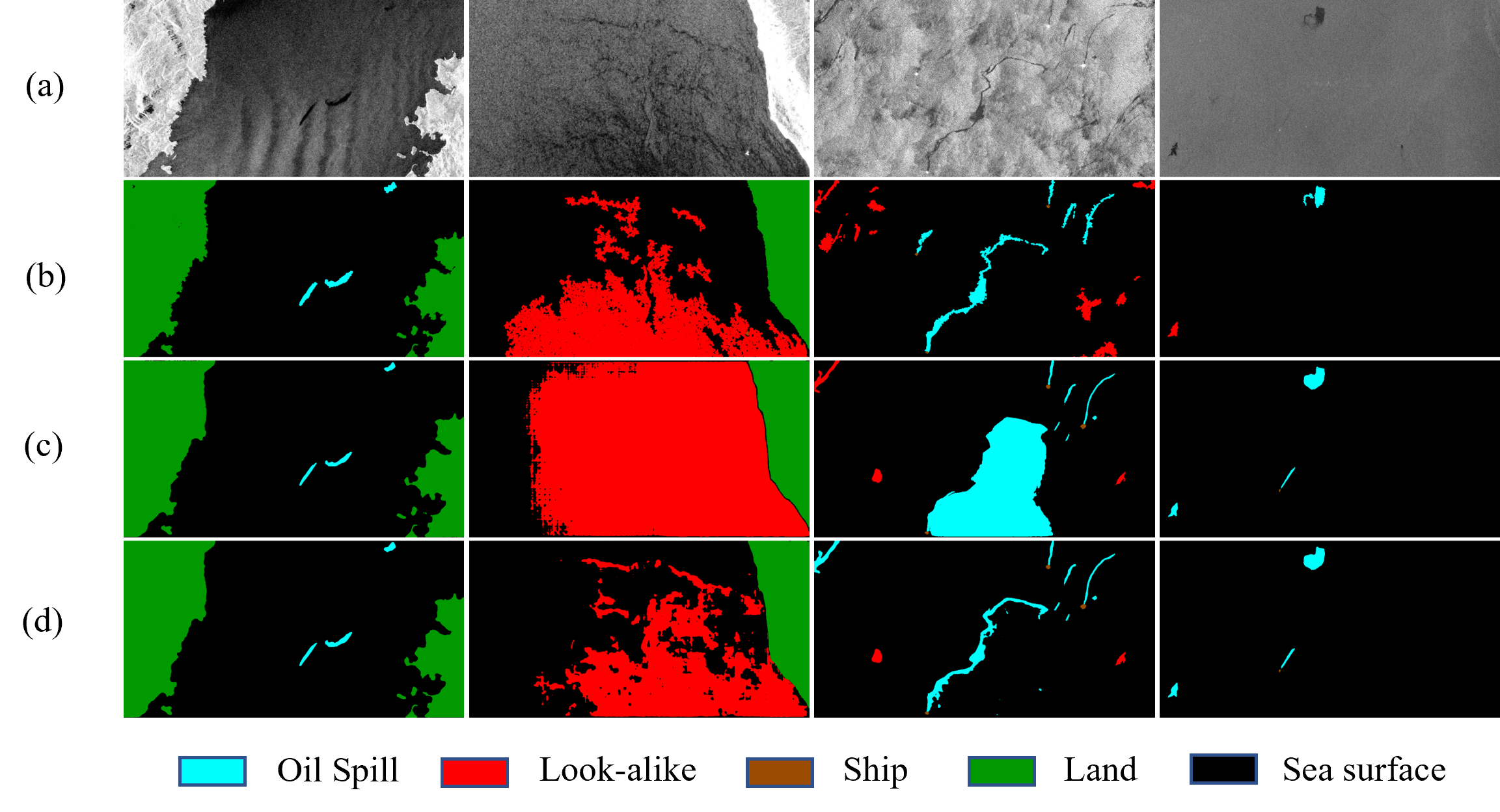}
  \caption{Qualitative examples of YOLOv8-SAM and SAM-OIL. (a) SAR Images, (b) Ground Truth, (c) YOLOv8-SAM, (d) SAM-OIL.}
  \label{Fig4}
\end{figure}

Fig. \ref{Fig4} shows that YOLOv8-SAM and SAM-OIL perform well in detecting elongated and small dark areas with simple surroundings, as seen in column 1. The integration of OMF and Adapter helps SAM-OIL segment objects with fuzzy boundaries more accurately, as shown in columns 2 and 3. However, small-sized look-alikes, particularly in the top left of column 3 and the bottom left of column 4, are sometimes misclassified as oil spills by the SAM-OIL, highlighting the challenges of using SAR imagery alone for oil spill detection.

\section{Conclusion} \label{conclusion}
We proposed a composite framework, SAM-OIL, which integrates YOLOv8 with the SAM, incorporating OMF and Adapter modules. To the best of our knowledge, this is the first time the SAM has been applied in oil spill detection. Compared with existing methods, SAM-OIL achieves the highest mIoU on the M4D dataset with 69.52\%, demonstrating the significant potential of the SAM-based composite framework in oil spill detection. Furthermore, the SAM-OIL version without the adapter module, YOLOv8-SAM+OMF, achieved 65.82\% mIoU, indicating the framework's ability on reducing the dependence on finely annotated samples.

Qualitative analysis shows that small-sized look-alikes are often confused with oil spills, suggesting the limitations in relying solely on SAR satellite imagery for detection and the need of ancillary data (e.g., oil platform locations, ship traffic)\cite{alpers2017oil}. Within the SAM-OIL framework, the object detector significantly contributes to the accuracy improvement. Hence, prioritizing object detectors could further improve detection performance in the future.

\balance
\bibliographystyle{IEEEtran}

\bibliography{sam-oil} 

\end{document}